%% file: main.tex
\documentclass[letterpaper,10pt,conference]{ieeeconf}

\IEEEoverridecommandlockouts %
\input{macros.tex}

\title{\LARGE \bf MOTLEE: Distributed Mobile Multi-Object Tracking\\%
with Localization Error Elimination}

\author{Mason B. Peterson*, Parker C. Lusk*, Jonathan P. How%
	\thanks{The authors are with the Department of Aeronautics and Astronautics, Massachusetts Institute of Technology.
	    {\texttt{\{masonbp, plusk, jhow\}@mit.edu.}} *Equal contribution.}
    \thanks{Work is supported by Ford Motor Company and ONR.}
}%

\begin{document}

\maketitle
\thispagestyle{plain}
\pagestyle{plain}

\begin{abstract} 
\input{paper/abstract}
\end{abstract}

\input{paper/01_intro}

\input{paper/02_related_work}

\input{paper/03_tracking_approach}

\input{paper/04_frame_alignment}

\input{paper/05_experiments}

\input{paper/06_conclusion}

\balance %

\bibliographystyle{IEEEtran}
\bibliography{refs}

\end{document}

%% file: macros.tex
\usepackage{amsmath} %
\usepackage{amssymb} %
\usepackage{amsfonts}
\usepackage[sort,compress]{cite}
\usepackage[usenames,dvipsnames]{xcolor}
\usepackage{graphicx}
\usepackage{tabularx}
\usepackage{multirow}
\usepackage{mathtools}
\usepackage{esvect} %
\usepackage[]{siunitx}
\usepackage{caption}
\usepackage[pdftex,pdfstartview={FitV},pdfpagelayout={TwoColumnLeft},bookmarksopen=true,plainpages=false,colorlinks=true,linkcolor=black,citecolor=black,urlcolor=black,filecolor=black ,pagebackref=false,hypertexnames=false,plainpages=false,pdfpagelabels]{hyperref}
\usepackage[T1]{fontenc} %
\usepackage{mathtools, cuted} %
\usepackage{bm}
\usepackage{xspace}
\usepackage{arydshln}

\usepackage{etoolbox}
\usepackage[normalem]{ulem}
\robustify\bfseries
\robustify\uline

\newcolumntype{H}{>{\setbox0=\hbox\bgroup}c<{\egroup}@{}}

\usepackage{enumerate}
\usepackage{balance} %
\usepackage{booktabs} %
\usepackage[export]{adjustbox} %

\newcolumntype{x}[1]{%
>{\raggedleft\hspace{0pt}}p{#1}}%

\definecolor{scan1blue}{HTML}{0072bd}
\definecolor{scan2red}{HTML}{d95319}

\usepackage{algorithm}
\usepackage[noend]{algpseudocode} %
\definecolor{commentclr}{RGB}{34, 139, 34}

\usepackage{tikz}
\usetikzlibrary{plotmarks}

\usepackage{subcaption}
\DeclareCaptionLabelSeparator{periodspace}{.\quad}
\usepackage{amsthm}
\usepackage{thmtools, thm-restate}

\theoremstyle{definition}

\makeatletter
\newcommand{\newreptheorem}[2]{%
\newtheorem*{rep@#1}{\rep@title}%
\newenvironment{rep#1}[1]{%
 \def\rep@title{#2 \ref*{##1}}%
 \begin{rep@#1}}%
 {\end{rep@#1}}}
\makeatother

\usepackage{letltxmacro}
\LetLtxMacro\orgvdots\vdots
\LetLtxMacro\orgddots\ddots

\makeatletter
\DeclareRobustCommand\vdots{%
  \mathpalette\@vdots{}%
}
\newcommand*{\@vdots}[2]{%
  \sbox0{$#1\cdotp\cdotp\cdotp\m@th$}%
  \sbox2{$#1.\m@th$}%
  \vbox{%
    \dimen@=\wd0 %
    \advance\dimen@ -3\ht2 %
    \kern.5\dimen@
    \dimen@=\wd2 %
    \advance\dimen@ -\ht2 %
    \dimen2=\wd0 %
    \advance\dimen2 -\dimen@
    \vbox to \dimen2{%
      \offinterlineskip
      \copy2 \vfill\copy2 \vfill\copy2 %
    }%
  }%
}
\DeclareRobustCommand\ddots{%
  \mathinner{%
    \mathpalette\@ddots{}%
    \mkern\thinmuskip
  }%
}
\newcommand*{\@ddots}[2]{%
  \sbox0{$#1\cdotp\cdotp\cdotp\m@th$}%
  \sbox2{$#1.\m@th$}%
  \vbox{%
    \dimen@=\wd0 %
    \advance\dimen@ -3\ht2 %
    \kern.5\dimen@
    \dimen@=\wd2 %
    \advance\dimen@ -\ht2 %
    \dimen2=\wd0 %
    \advance\dimen2 -\dimen@
    \vbox to \dimen2{%
      \offinterlineskip
      \hbox{$#1\mathpunct{.}\m@th$}%
      \vfill
      \hbox{$#1\mathpunct{\kern\wd2}\mathpunct{.}\m@th$}%
      \vfill
      \hbox{$#1\mathpunct{\kern\wd2}\mathpunct{\kern\wd2}\mathpunct{.}\m@th$}%
    }%
  }%
}
\makeatother

\let\oldnl\nl%
\newcommand{\nonl}{\renewcommand{\nl}{\let\nl\oldnl}}%
\makeatother

\def\sl{\mathcal{L}}

\newcommand{\SE}[1]{\ensuremath{\mathrm{SE}(#1)}\xspace}

\newcommand\eqdef{\mathrel{\overset{\makebox[0pt]{\mbox{\normalfont\tiny def}}}{=}}}

\graphicspath{{figures/}}

\usepackage{svg}
\svgpath{{figures/}}

%% file: paper/abstract.tex
We present MOTLEE, a distributed mobile multi-object tracking algorithm that enables a team of robots to collaboratively track moving objects in the presence of localization error.
Existing approaches to distributed tracking make limiting assumptions regarding the relative spatial relationship of sensors, including assuming a static sensor network or that perfect localization is available.
Instead, we develop an algorithm based on the Kalman-Consensus filter for distributed tracking that properly leverages localization uncertainty in collaborative tracking.
Further, our method allows the team to maintain an accurate understanding of dynamic objects in the environment by realigning robot frames and incorporating frame alignment uncertainty into our object tracking formulation.
We evaluate our method in hardware on a team of three mobile ground robots tracking four people.
Compared to previous works that do not account for localization error, we show that MOTLEE is resilient to localization uncertainties, enabling accurate tracking in distributed, dynamic settings with mobile tracking sensors.

%% file: paper/01_intro.tex
\section{Introduction}\label{sec:intro}

Reliable deployment of autonomous vehicles (AVs) requires that they have an accurate and up-to-date knowledge of the environment.
Dynamic objects require particular care so that an AV can safely perform decision making in the presence of pedestrians and other moving agents.
Multiple object tracking (MOT) provides a framework to allow an AV to detect, track, and forecast future motion of dynamic objects~\cite{bewley2016sort,chiu2021probabilistic}.
However, field of view limitations can degrade an AV's ability to sense the environment, and although state-of-the-art methods have shown increased accuracy and robustness by fusing multiple modality sensors that are rigidly attached to the AV~\cite{zhang2019robust,kim2021eagermot,wang2022detr3d}, factors such as occlusion and sensing range can limit the effective field of view of MOT algorithms.
These limitations have led to research on multi-view MOT~\cite{fleuret2007multicamera,chavdarova2018wildtrack,casao2021distributed,nguyen2022multi}, which allows AV teams to collaboratively track dynamic objects~\cite{shorinwa2020distributed}.

Multi-view MOT, which uses multiple sensors networked together for tracking, has the benefit of an increased tracking coverage area, but requires accurate knowledge of the spatial relationships (i.e., the localization) between sensors so that measurements can be correctly interpreted.
Existing methods typically assume either a static, calibrated sensor network~\cite{fleuret2007multicamera,sandell2008distributed,chavdarova2018wildtrack,casao2021distributed} or moving sensors within a known map and with perfect localization~\cite{shorinwa2020distributed,kroeger2014multi,dames2015autonomous,dames2020distributed}.
However, in practice these assumptions are violated due to mis-calibration, environmental effects (e.g., vibration, temperature), or localization uncertainty, causing sensors' local coordinate frames to become misaligned.
Misaligned local frames can have disastrous consequences in multi-sensor estimation problems because data shared between sensors may be inconsistent, leading to incorrect data association and noisy signal recovery.

In the context of multi-view MOT for AVs, these frame misalignment issues can cause the team to generate false-positive tracks (i.e., tracks for objects that are not really there), which corrupt AVs' understanding of the environment and thus negate the value of shared knowledge.
Localization, and therefore frame alignment, will never be perfect; therefore, the multi-view MOT fusion process must understand this uncertainty.
Object measurement noise must properly reflect both an AV's own localization uncertainty and the uncertainty of frame alignment when measurements are sent to neighboring AVs. 
Thus, AVs with more localization uncertainty can have their measurements appropriately down-weighted relative to other AVs' information.
Otherwise, the estimation process may become over-confident about the wrong answer (i.e., filter inconsistency).
Because we are interested in distributed MOT, which provides scalability, redundancy, and communication resilience~\cite{liggins1997distributed,taj2011distributed}, frame alignment issues cannot be solved by a central authority and must be handled by each AV.

\begin{figure}[t]
    \centering
    \includegraphics[width=\columnwidth]{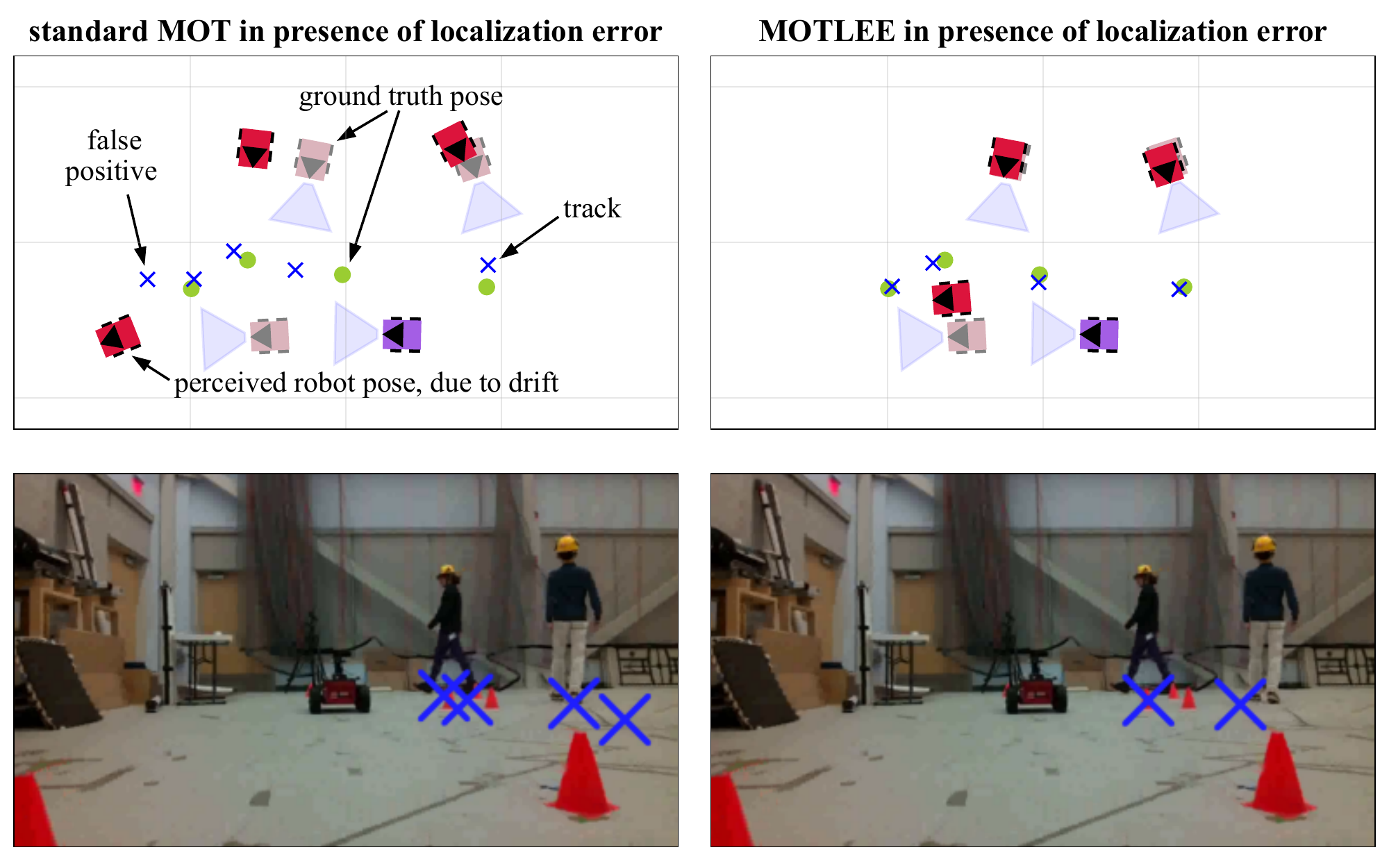}
    \caption{
    Hardware experiment comparing MOTLEE (right) with standard MOT, which does not account for localization uncertainty (left).
    Perceived and true poses of robots are shown from the perspective of the purple robot.
    MOTLEE allows robots to understand spatial relationships with each other, leading to objects being tracked accurately.
    Without MOTLEE, robots cannot account for localization error, leading to a poor understanding of the environment.
    See supplementary video for further results.
}
    \label{fig:motlee_fig1}
\end{figure}

We address the issue of localization uncertainty in distributed mobile MOT by providing a decentralized mechanism for online frame alignment between pairs of robots.
Further, we properly fuse measurements according to each robot's localization uncertainty and relative frame alignment uncertainty.
Our method, called MOTLEE (MOT with Localization Error Elimination) performs distributed tracking in the ground plane using the Kalman-Consensus filter (KCF)~\cite{olfati2007distributed,olfati2009kalman} and corrects frame misalignment by using both static landmarks and dynamic objects that are co-visible between AVs.
We demonstrate the performance of MOTLEE on a team of mobile ground robots tracking a group of pedestrians, shown in Fig.~\ref{fig:motlee_fig1}.
We experimentally show that even a small amount of localization error (\SI{0.25}{\m}, \SI{2}{\deg}) causes existing methods to break down, while our method is able to maintain high accuracy.
In summary, our contributions include:
\begin{enumerate}
    \item A decentralized algorithm for compensating for localization error between pairs of robots, relaxing assumptions made by state-of-the-art distributed MOT systems to enable distributed \emph{mobile} MOT.
    \item A distributed MOT system that accounts for measurement uncertainty due to localization and frame alignment errors. Properly modelling the noise covariance prevents failed data association and over-confident estimation.
    \item Hardware demonstration of our distributed MOTLEE method with a team of three ground robots tracking four pedestrians. We show that our system, in the presence of localization error, is able to achieve a tracking accuracy very close to the ideal tracking accuracy of a system with perfect localization.
\end{enumerate}

%% file: paper/02_related_work.tex
\section{Related Work}\label{sec:relatedwork}
Tracking multiple objects using sensor observations is a well-studied task spanning multiple communities~\cite{blackman1986multiple,bar1995multitarget,luo2021motlitreview}.
State-of-the-art systems employ \emph{tracking-by-detection}, wherein object detections are associated across time and then fused to estimate object trajectories.
Significant research has been devoted to improving these systems by focusing on improving data association~\cite{zhang2008global,yang2017hybrid,weng2020gnn3dmot}, fusing multiple modalities~\cite{kim2021eagermot,zhang2019robust,wang2022deepfusionmot}, or using multiple sensors~\cite{fleuret2007multicamera,ristani2018motreid,tesfaye2019multi}.
In the following discussion, we focus on works most relevant to localization uncertainty in the context of MOT.

\subsection{Multi-view MOT with Known Localization}
Multi-view MOT, sometimes referred to as multi-target multi-camera tracking (MTMCT), is most commonly performed with static cameras having known spatial relationships~\cite{fleuret2007multicamera,ristani2018motreid,casao2021distributed,soto2009distributed,sandell2008distributed,nguyen2021distributed}, e.g., through extrinsic calibration.
These systems may be centralized~\cite{fleuret2007multicamera,ristani2018motreid} or distributed~\cite{casao2021distributed,soto2009distributed,sandell2008distributed,nguyen2021distributed}.
In centralized systems, each sensor node sends information to a central fusion center, where all data is processed together in a common coordinate frame.
Centralized methods typically achieve high tracking accuracy, but have the main disadvantages of using excessive communication and having a single point of failure.
Distributed methods alleviate these disadvantages, but still typically assume that cameras are static and never perturbed, often requiring an offline calibration step beforehand.
Pan-tilt-zoom (PTZ) cameras are static in position, but may rotate; however, these parameters are typically communicated to the other cameras so that coordinate frames align~\cite{soto2009distributed}.

Mobile multi-view MOT systems also require knowledge of spatial relationships, which change over time.
Kroeger et al.~\cite{kroeger2014multi} performed centralized multi-view tracking using mobile cameras by localizing cameras in a dense map previously reconstructed using structure from motion.
Ong et al.~\cite{ong2006decentralised} proposed a decentralized particle filtering approach and tested it using multiple flight vehicles, but assume perfect localization in a common global frame (e.g., using GPS).
Many mobile multi-view MOT systems arise in multirobot tasks, such as multi-target mapping~\cite{dames2015autonomous}, multi-target search and tracking~\cite{dames2020distributed,zahroof2022multi}, and multi-target coverage control~\cite{chen2020collision}.
Shorinwa et al.~\cite{shorinwa2020distributed} presented a distributed MOT algorithm based on consensus ADMM to enable collaborative tracking among AVs, but made strong assumptions about known data association from each measurement and perfect AV localization.
Before these algorithms can be deployed on hardware in real-world scenarios, the perfect localization assumption must be relaxed.

\subsection{Single-view MOT with Unknown Localization}
More effort has been made in relaxing the known localization assumption for the single-view (i.e., single robot) MOT case and is closely related to simultaneous localization and mapping methods.
Wang et al.~\cite{wang2002simultaneous,wang2003online,wang2007slammot} first formulated the joint SLAM and MOT problem.
In \cite{wang2007slammot}, the authors presented a maximum a posteriori (MAP) formulation of SLAMMOT using a Bayes network and implemented an approximate version that decouples the SLAM and MOT problem into two separate estimators for static and dynamic objects.
Similarly, Choi et al.~\cite{choi2012general} presented a sampling-based method for simultaneously estimating object tracks and the camera's ego-motion by leveraging static features in the scene.
Recent works continue to build on these ideas for single-view MOT by leveraging factor graphs for 3D tracking and ego-motion estimation~\cite{tian2022dlslot} or using MOT as a mechanism to improve visual SLAM accuracy~\cite{zhang2022motslam}.

\subsection{Multi-view MOT with Unknown Localization}
Few works have considered localization uncertainty in multi-view MOT. 
Lin et al.~\cite{lin2004exact} showed that local estimates of tracked objects could be used to estimate dynamic biases of collaborative static sensors.
Their approach is specific to drift characteristic of range and bearing sensors, with which they create a constrained problem and solve for biases in each sensor.
However, they do not not address spatial localization uncertainty experienced by mobile sensors, and their simulation results are limited, involving tracking a known number of targets without occlusions.
Moratuwage et al.~\cite{moratuwage2013collaborative} presented a collaborative SLAM and tracking method based on random finite sets (RFS) and the probabilistic hypothesis density (PHD) filter.
However, their algorithm is centralized and so requires robots to communicate all their data to a central machine.
Ahmad et al.~\cite{ahmad2013cooperative} formulated a joint, collaborative MOT and localization problem as a pose graph optimization for robotic soccer.
They make the limiting assumption of known data association of measurements with static landmarks and tracked dynamic objects. 
Chen and Dames~\cite{chen2020collision} recently proposed a distributed MOT algorithm, also based on RFS and the PHD filter.
Their work introduces the notion of convex uncertainty Voronoi cells, which describes the area each robot is assigned to patrol.
Because targets may cross cell boundaries, additional communication is required to transfer ownership of targets and to update the cells.
Simulation results of their method are provided, however, the localization uncertainty everywhere along the robot's path is assumed to be constant, which is not realistic.

%% file: paper/03_tracking_approach.tex
\section{Distributed Mobile MOT}\label{sec:mot}

\begin{figure*}[t]
    \centering
    \includeinkscape[pretex=\footnotesize,width=\textwidth]{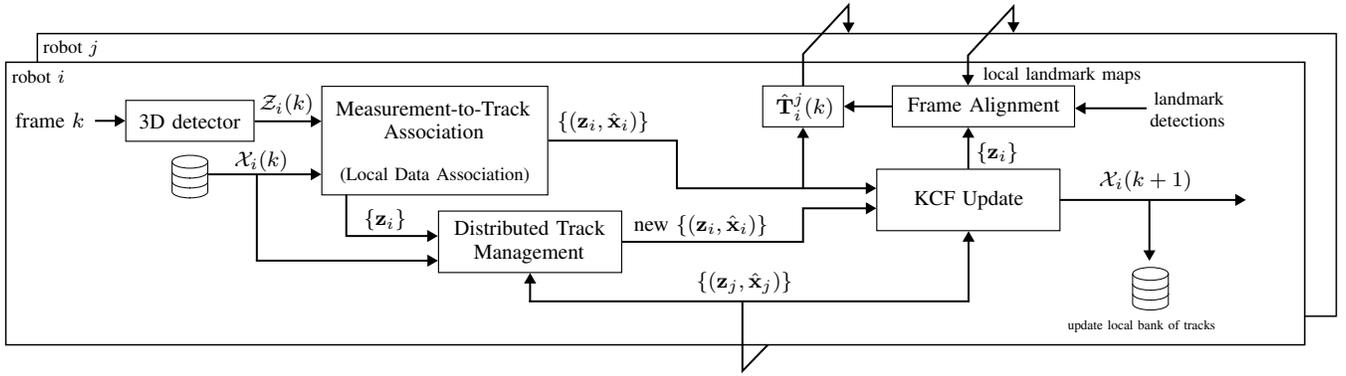}
    \caption{
    MOTLEE system architecture.
    At each camera frame, robot $i$ extracts detections which are then associated with the robot's local set of tracks, $\mathcal{X}_i(k)$, in the local data association step.
    Robot $i$ then sends the measurements to robot $j \in \mathcal{N}_i$ after the measurements have been transformed with $\hat{\mathbf{T}}_i^j(k)$ into robot $j$'s local frame.
    The KCF then collects all measurements for a given track from robot $i$'s neighbors and fuses them in robot $i$'s local frame.
    Correctly transforming into other robots' local frames is crucial for tracking accuracy, and so a frame alignment algorithm (see Section~\ref{sec:frame-alignment}) is used to periodically estimate $\hat{\mathbf{T}}_i^j(k)$.
    Frame alignment can be performed with either associated detections of moving objects or a map of static landmarks as inputs.
    }
    \label{fig:block-diagram}
\end{figure*}

We formulate distributed MOT within the Kalman-Consensus filter (KCF) framework~\cite{olfati2007distributed,olfati2009kalman} and adopt the distributed track manager methodology of Casao~et~al.~\cite{casao2021distributed}.
Consider a network of $N$ robots with the communication graph $\mathcal{G}$, where each vertex in the graph is a robot.
We assume that the graph is connected (i.e., a path exists between any two vertices) and denote the one-hop neighbors of robot $i$ as $\mathcal{N}_i$.
Quantities associated with the $i$-th robot are denoted with subscript $i$.
An unknown number of moving objects are observed by the robot team, with each object modeled by a discrete-time linear dynamic system
\begin{equation}\label{eq:ltisys}
    \mathbf{x}(k+1) = \mathbf{A}\mathbf{x}(k) + \mathbf{w}(k), \quad \mathbf{z}(k) = \mathbf{H}\mathbf{x}(k) + \mathbf{v}(k),
\end{equation}
where $\mathbf{A}$ and $\mathbf{H}$ are the transition and measurement matrices, respectively, and $\mathbf{w}(k)\sim\mathcal{N}(0,\mathbf{Q}(k))$ and $\mathbf{v}(k)\sim\mathcal{N}(0,\mathbf{R}(k))$ are zero-mean independent Gaussian process and measurement noise with covariance matrices $\mathbf{Q}$ and $\mathbf{R}$, respectively.
We define a \emph{track} as the estimate $\hat{\mathbf{x}}(k)$ corresponding to measurements of an object.
The \mbox{$i$-th} robot maintains its own local bank of tracks, denoted ${\mathcal{X}_i(k)=\{\hat{\mathbf{x}}_i^t(k)\}_{t=1}^{|\mathcal{X}_i(k)|}}$.
The $t$-th track has an associated error covariance $\mathbf{P}^t_i(k)$, along with other variables including track lifetime and number of missed detections, which are used for track management.
Ideally, each track corresponds to a single object, but this may not always be the case due to noise, partial measurements, and clutter.

Typical single-view MOT systems consist of the following stages: 1) measurement-to-track data association, 2) data fusion, and 3) track management (initialization and deletion).
In distributed MOT, measurement-to-track data association is performed locally, while data fusion and track management is performed in a distributed manner.
Additionally, distributed track-to-track data association is required to ensure the team maintains a consistent understanding when new tracks are shared.
Furthermore, because the robots are mobile and have localization uncertainty, robots must transform measurements and their covariances into their neighbors' coordinate frames when sharing data.
These stages are depicted in the block diagram in Fig.~\ref{fig:block-diagram} and discussed in the following subsections.

\subsection{Localization Uncertainty}\label{sub:uncertainty}

Because the robots are mobile, they must localize themselves with respect to each other so that measurements from the $i$-th robot can be used by its neighbors.
We denote the pose of the \mbox{$i$-th} robot at time $k$ as $\mathbf{T}_i(k)\in\SE{d}$, where $d=2$ or $3$, depending on the tracking scenario.
Note that each robot's local frame is defined as its initial frame, and $\mathbf{T}_i(k)$ is expressed with respect to robot $i$'s local frame.
Because of localization errors (e.g., odometric drift), each robot only has access to an estimate of its pose, denoted $\hat{\mathbf{T}}_i(k)$, and its associated pose uncertainty $\bm{\Sigma}_i(k)$.
Tracks $\mathcal{X}_i(k)$ are expressed in the robot's local frame and inter-robot frame alignments $\mathbf{T}^j_i(k)$ are required for data fusion.

Consensus on geometric quantities requires a shared coordinate frame.
In particular, as robots communicate their local measurements to their neighbors, a pair of robots $i$ and $j$ must be able to transform measurements into the other's frame.
Supposing there exists a previous transform $\mathbf{T}^j_i(k_0)$ between their local frames, measurements can easily be transformed across robots.
This previous transform can be found if robots are localizing within a shared reference map or if the robots start with known relative poses.
However, in the presence of localization uncertainty, drift causes frame misalignment between pairs of robots.

\begin{figure}[t]
    \centering
    \includeinkscape[pretex=\footnotesize,width=\columnwidth]{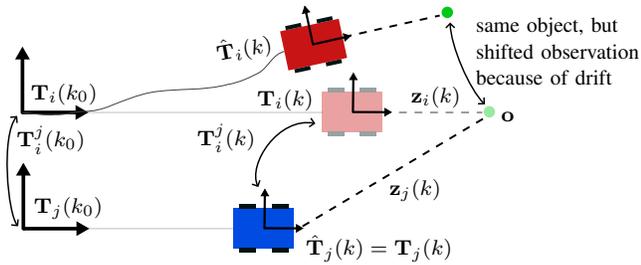}
    \caption{
    Effects of localization error (i.e., odometry drift) on communication about measurements.
    Robot $i$ (red) makes an observation $\mathbf{z}_i(k)$, but because of the drift in its current pose, the object appears shifted in its local frame.
    Robot $j$ (blue) makes an observation $\mathbf{z}_j(k)$, but has no drift in its current pose, so the object is correctly mapped in its local frame.
    If the outdated relative transform $\mathbf{T}^j_i(k_0)$ is used, drift from robot $i$ causes the object measurements to be misaligned.
    If a new $\mathbf{T}^j_i(k)$ can be estimated, the effects of localization error are eliminated and $\mathbf{z}_i(k),\mathbf{z}_j(k)$ align.
    }
    \label{fig:loc-err}
\end{figure}

Consider the scenario in Fig.~\ref{fig:loc-err}, where, for simplicity and without loss of generality, robot $i$ has accumulated drift since time $k_0$, but robot $j$ has not.
At time $k$, both robots make a measurement of a single object, denoted $\mathbf{o}$.
Robot $i$ receives a measurement $\mathbf{z}_i(k)$ of the object, but due to drift, it is incorrectly mapped into its local frame.
Since robot $j$ maps the object perfectly (no drift), the expression of $\mathbf{z}_i(k)$ in the frame $\hat{\mathbf{T}}_j(k)$ is far from its measurement $\mathbf{z}_j(k)$ and so fusing these measurements results in a poor estimate of the position of $\mathbf{o}$.
Instead, if $\mathbf{T}^j_i(k)$ was found and used, the effects of drift would be removed and $\mathbf{z}_i(k)$, $\mathbf{z}_j(k)$ would be perfectly aligned.
This example highlights the important role of $\mathbf{T}^j_i(k)$---it allows robots to communicate coherently about measurements received in each other's local frame.
In Section~\ref{sec:frame-alignment}, we will present methods for providing the estimate $\hat{\mathbf{T}}^j_i(k)\sim\mathcal{N}(\mathbf{T}^j_i(k),\mathbf{\Sigma}^j_i(k))$.

\subsection{Localization Uncertainty Propagation}\label{sub:uncertainty_propagation}

Data association and fusion processes assume valid noise statistics.
If the noise covariance is not representative of the true uncertainty, then data association can fail and fusion can incorrectly incorporate information.
To account for localization uncertainty $\bm{\Sigma}_i$ of robot $i$, measurements $\mathbf{z}_i\sim\mathcal{N}(\bar{\mathbf{z}}_i, \mathbf{R}_i)$ must be properly combined when transforming $\mathbf{z}_i$ into the local frame of robot $i$.
Following Smith, Self, and Cheeseman~(SSC)~\cite{Smith1990}, we parameterize the transformation $\hat{\mathbf{T}}_i$ using Euler angles and the pose composition (head-to-tail) combines the covariances of $\bm{\Sigma}_i$ and $\mathbf{R}_i$ as
\begin{equation}\label{eq:transform-cov-into-local}
    \tilde{\mathbf{R}}_i \approx \mathbf{F} \bm{\Sigma}_i \mathbf{F}^\top + \mathbf{G} \mathbf{R}_i \mathbf{G}^\top,
\end{equation}
where $\mathbf{F}$ and $\mathbf{G}$ are Jacobians encoding the transformation $\hat{\mathbf{T}}_i$.
This covariance, associated with the transformed measurement $\tilde{\mathbf{z}}_i=\hat{\mathbf{T}}_i\mathbf{z}_i$, are then used for making local data association decisions (see Section~\ref{sub:lda}).
Similarly, when transforming a measurement $\mathbf{z}_i$ for a neighboring robot $j$ using the relative transform $\hat{\mathbf{T}}^j_i$, the measurement and covariance are transformed using the tail-to-tail method
\begin{align}
    \tilde{\mathbf{z}}_j &= \hat{\mathbf{T}}^j_i \mathbf{z}_i \label{eq:transform-meas-into-j} \\
    \tilde{\mathbf{R}}_j &\approx \mathbf{J} \tilde{\mathbf{R}}_i \mathbf{J}^\top + \bm{\Sigma}^j_i, \label{eq:transform-cov-into-j}
\end{align}
where $\mathbf{J}$ is the Jacobian encoding the frame alignment $\hat{\mathbf{T}}_i^j(k)$.

\subsection{Kalman-Consensus Filter}\label{sub:kcf}

The KCF is a distributed estimation algorithm with a peer-to-peer architecture~\cite{sandell2008distributed}, and like existing works~\cite{casao2021distributed,sandell2008distributed}, we assume that sensor nodes are synchronized; however, this assumption can be relaxed as in the event-triggered KCF of~\cite{li2016event}.
At each timestep $k$, each of the $N$ robots makes a set of observations $\mathcal{Z}_i(k)=\{\mathbf{z}_i^m(k)\}_{m=1}^{M_i},i=1,\dots,N$.
After local data association is solved (see Section~\ref{sub:lda}), new local measurements are fused with their associated track in the robot's local bank $\mathcal{X}_i(k)$.
Because each robot makes noisy measurements in its own local frame, robots' local estimates of a given track will differ.
However, consensus is enabled through frame alignment estimates $\hat{\mathbf{T}}^j_i(k)$.
Using the KCF allows the robots to reach consensus about track estimates via transformed local frames and enables sharing of measurements to produce more accurate estimates.

For clarity of exposition, we consider a single KCF iteration from the point of view of robot $i$ where its local copy of track $\hat{\mathbf{x}}_i^t$ is associated with the local measurement $\mathbf{z}_i^m$ (see Section~\ref{sub:lda}).
For brevity, we drop the superscripts $t$ and $m$.
The KCF performs consensus filtering in information form, thus, measurements are first transformed according to 
\begin{equation}
    \mathbf{u}_j = \mathbf{H}^\top \tilde{\mathbf{R}}_j^{-1}\tilde{\mathbf{z}}_j, \quad \mathbf{U}_j = \mathbf{H}^\top \tilde{\mathbf{R}}_j^{-1}\mathbf{H},
\end{equation}
where the measurement and its covariance are first transformed into the receiving neighbor's frame according to \eqref{eq:transform-meas-into-j} and \eqref{eq:transform-cov-into-j}.
The $i$-th robot then broadcasts the message $(\hat{\mathbf{x}}_j^+, \mathbf{u}_j, \mathbf{U}_j, \texttt{ID}_j)$ to its neighbors.
Once robot $i$ receives this message from each of its neighbors, the information is aggregated as
\begin{equation}
    \mathbf{y}_i = \sum_{j\in\mathcal{N}_i\cup i} \mathbf{u}_j, \qquad \mathbf{Y}_i = \sum_{j\in\mathcal{N}_i\cup i} \mathbf{U}_j.
\end{equation}
Each track is then updated with the fused information from all robots using the KCF update,
\begin{gather}
\begin{aligned}
    \hat{\mathbf{x}}_i(k+1) &= \hat{\mathbf{x}}_i^+(k) + \mathbf{M}_i(k)\bigl[\mathbf{y}_i(k) - \mathbf{Y}_i(k)\hat{\mathbf{x}}_i^+(k)\bigr] \\
    &\quad+ \frac{\mathbf{M}_i(k)}{1+\|\mathbf{M}_i(k)\|}\sum_{j\in\mathcal{N}_i} \bigl(\hat{\mathbf{x}}_j^+(k) - \hat{\mathbf{x}}_i^+(k)\bigr),
\end{aligned}
\end{gather}
where $\mathbf{M}_i(k) = (\mathbf{P}_i(k)^{-1} + \mathbf{Y}_i(k))^{-1}$ is the Kalman gain in information form and $\mathbf{P}_i$ is the error covariance of the track state.
Finally, the covariance matrix is updated according to $\mathbf{P}(k+1)=\mathbf{A}\mathbf{M}_i(k)\mathbf{A}^\top + \mathbf{Q}_i(k)$ and $\hat{\mathbf{x}}_i^+ = \mathbf{A}\hat{\mathbf{x}}_i$.

\subsection{Local Data Association}\label{sub:lda}

Local data association (LDA) assesses which track best explains each measurement.
We solve LDA using the global nearest neighbor (GNN) approach~\cite{bar1995multitarget}, which formulates the data association problem as a linear assignment problem using a matching cost given by the Mahalanobis distance
\begin{equation}
    d(\tilde{\mathbf{z}}_i^m, \hat{\mathbf{x}}_i^{t+}) = \|\tilde{\mathbf{z}}_i^m - \mathbf{H}\hat{\mathbf{x}}_i^{t+}\|^2_{(\mathbf{S}_i^t)^{-1}},
\end{equation}
where $\mathbf{S}_i^t=\mathbf{H}\mathbf{P}_i^{t+}\mathbf{H}^\top + \tilde{\mathbf{R}}_i^m$ is the innovation covariance.
Note that we use the transformed measurement and covariance according to \eqref{eq:transform-cov-into-local}.
We use a gating region defined by $\tau_\text{gate}$, so that for $d>\tau_\text{gate}$ a measurement cannot be assigned to the track.
GNN can be solved exactly and in polynomial time using the Hungarian algorithm~\cite{kuhn1955hungarian}.

\subsection{Track Management and Distributed Data Association}\label{sub:dda}

Track management is an important component of MOT systems.
The distributed case involves robots making decisions about when to create new tracks and when to delete an existing track that has not recently received any new measurements.
See Casao et al.~\cite{casao2021distributed} for an overview of performing distributed track management.

%% file: paper/04_frame_alignment.tex
\section{Frame Realignment Algorithm}\label{sec:frame-alignment}

For robot $i$ to communicate with robot $j$ about measurements in robot $j$'s own local frame, robots must have an estimate of the transformation between frames, $\hat{\mathbf{T}}_i^j(k)$.
A poor estimate $\hat{\mathbf{T}}_i^j(k)$ can prohibit robot $i$ from performing correct data association between its own measurement and the corrupted track information from robot $j$, so it is crucial that  $\hat{\mathbf{T}}_i^j(k)$ is an accurate estimate of the robots' true relative transformation. 
In this section, we will present a distributed method for eliminating error between robot frames using local maps of static landmarks and past dynamic object detections.

\subsection{Realignment with Static Landmarks}

The key concept is that frame alignment drift can be accounted for and eliminated by aligning objects observed by pairs of robots.
Because drift can accumulate and change quickly, our method aims to promptly correct for drift online by creating small, local maps of recently observed static landmarks.
With two local maps in hand, performing frame realignment with each neighbor involves three steps: first, associating landmarks between maps; second, applying a weighting to pairs of landmarks; and finally, performing point registration to compute $\hat{\mathbf{T}}^j_i(k)$.

To perform data association between two maps, we use iterative closest point (ICP) registration~\cite{besl1992method}.
As ICP is a local method that requires a good initial guess, we use the previous estimate $\hat{\mathbf{T}}_i^j(k-1)$ as a starting solution.
Weights are applied to the associations found by ICP using the following weighting function to prioritize recent detections over old detections,
\begin{equation}
    W(\ell_i, \ell_j) = ({\ell_i}{\ell_j})^{-1},
\end{equation}
where a larger weight corresponds to a greater influence in the point registration and $\ell_i$ and $\ell_j$ are the number of frames since the corresponding static landmarks were last detected by robots $i$ or $j$ respectively.
This is done to help account for drift that may have accumulated in older parts of the static local map.

The final step of the algorithm is to compute $\hat{\mathbf{T}}^j_i(k)$ using Arun's method~\cite{arun1987least} on the associated landmarks and respective weights. 
The transformation $\hat{\mathbf{T}}_i^j(k)$ should be applied to all outgoing detections sent to robot $j$ to place the measurements properly in robot $j$'s frame.

\subsection{Realignment with Tracked Objects}

When a large enough number of dynamic objects are co-visible, robots can use these already tracked objects and their measurements to perform frame alignment.
To determine whether dynamic object detections can be used to perform frame realignment, we define $\eta$ to be the number of concurrent, same-object detections between two agents and $\tau_\eta$ to be a threshold such that frame realignment is only performed with dynamic object detections when $\eta \geq \tau_\eta$.
If $\eta < \tau_\eta$, then, static landmark realignment should be performed.
Thus, with enough co-visibility for robots $i$ and $j$ to achieve a large $\eta$, determining the realignment transformation can be done without performing any additional mapping and by only exchanging information about tracked objects.

One important difference between aligning static landmarks and tracked object measurements is that measurements from tracked objects have already been associated during the KCF information exchange.
So, given two pairs of already-associated dynamic object measurements, $\mathcal{Z}_i$ and $\Tilde{\mathcal{Z}}_j$, we only need to find all pairs of detections $\mathbf{z}_i(k)$, $\Tilde{\mathbf{z}}_j(k)$ that occurred at the same time $k$ to perform realignment.

To realign frames with greater accuracy, we assign pairs of detections a weight that reflects the knowledge the agents have about the accuracy of the detection. 
We define the weighting function 
\begin{equation}
    W(\hat{\mathbf{x}}_i, \mathbf{z}_i, \Tilde{\mathbf{z}}_j) = ((\mathbf{H}\hat{\mathbf{x}}_i - \mathbf{z}_i)^{\top}(\mathbf{H}\hat{\mathbf{x}}_i - \Tilde{\mathbf{z}}_j))^{-1},
\end{equation}
which prioritizes aligning pairs of detections where both are consistent with the estimated state of the object, thus rejecting noisy detections and even detections that may have been associated incorrectly.

Once this step has been performed, a transformation is found using Arun's method for registration \cite{arun1987least}. 
Since aligned measurements $\mathbf{z}_i(k)$, $\Tilde{\mathbf{z}}_j(k)$ were already placed into robot $i$'s local frame, alignment of these measurements results in an intermediate correction transformation, $\mathbf{T}_\text{realign}$. So, to conclude frame realignment, $\hat{\mathbf{T}}_i^j(k)$ is updated with ${
    \hat{\mathbf{T}}_i^j(k) = \mathbf{T}_\text{realign} \hat{\mathbf{T}}_i^j(k - 1)}$.

One problem that may arise with this algorithm is that it assumes that frame alignment between two robots is accurate enough that robots can associate detections of common objects. 
This will not be the case under severe frame misalignment. 
To address this issue and to give robots the ability to make correct data associations when frame error is large, we make $\tau_\text{gate}$ (see Section~\ref{sub:lda}) reactive to the amount of detected frame misalignment, indicated by the magnitude of $\mathbf{T}_\text{realign}$.
The data association tolerance $\tau_\text{gate}$ is scaled up when frame realignment yields a large $\mathbf{T}_\text{realign}$ and $\tau_\text{gate}$ is returned back to its original value when frame realignment begins to yield a small $\mathbf{T}_\text{realign}$.

%% file: paper/05_experiments.tex
\section{Experiments}\label{sec:experiments}

We evaluate our distributed mobile MOT system using two self-collected datasets generated using a team of ground robots in a \SI[parse-numbers=false]{10\times10}{\meter} room.
Each robot was equipped with an Intel RealSense T265 Tracking Camera and an Intel RealSense L515 LiDAR Camera.
A VICON motion capture system was used to collect ground truth pose information of robots, moving objects (pedestrians), and static landmarks (cones).
For simplicity, we assume objects evolve according to \eqref{eq:ltisys} with a constant velocity motion model---thus the state of an object is defined to be ${\mathbf{x}(k)\eqdef\left[p_x,p_y,v_x,v_y\right]^\top\in\mathbb{R}^4}$.
Each robot makes noisy 2D position observations in the ground plane so that the measurement model is defined as ${\mathbf{z}(k)\eqdef\left[\tilde{p}_x,\tilde{p}_y\right]^\top\in\mathbb{R}^2}$, where object measurements are made by processing CenterTrack~\cite{zhou2020centertrack} trained on the JRDB dataset~\cite{martinmartin2023jrdb,anderson2022implementation} on T265 left fisheye images. 
Although CenterTrack additionally provides IDs associated with detections (i.e., the CenterTrack network attempts to infer data association), we find these to be too noisy for practical use. %

In addition to providing images used for person detection, the T265 stereo camera provides each robot with ego-motion estimation based on stereo visual odometry.
Although each robot begins with knowledge of relative frame alignments (e.g., by initializing $\hat{\mathbf{T}}_i^j(k_0) = \mathbf{T}_i^j(k_0)$), odometry drift quickly grows resulting in frame misalignments that cause failure of distributed data association and fusion of incoherent data in the KCF.
Using our MOTLEE algorithm, we address these challenges by performing frame alignment and appropriately incorporating odometry and frame alignment uncertainty into shared measurement uncertainty.

\begin{figure}[t]
    \centering
    \includegraphics[trim=0 1cm 0 1cm, clip,width=\columnwidth]{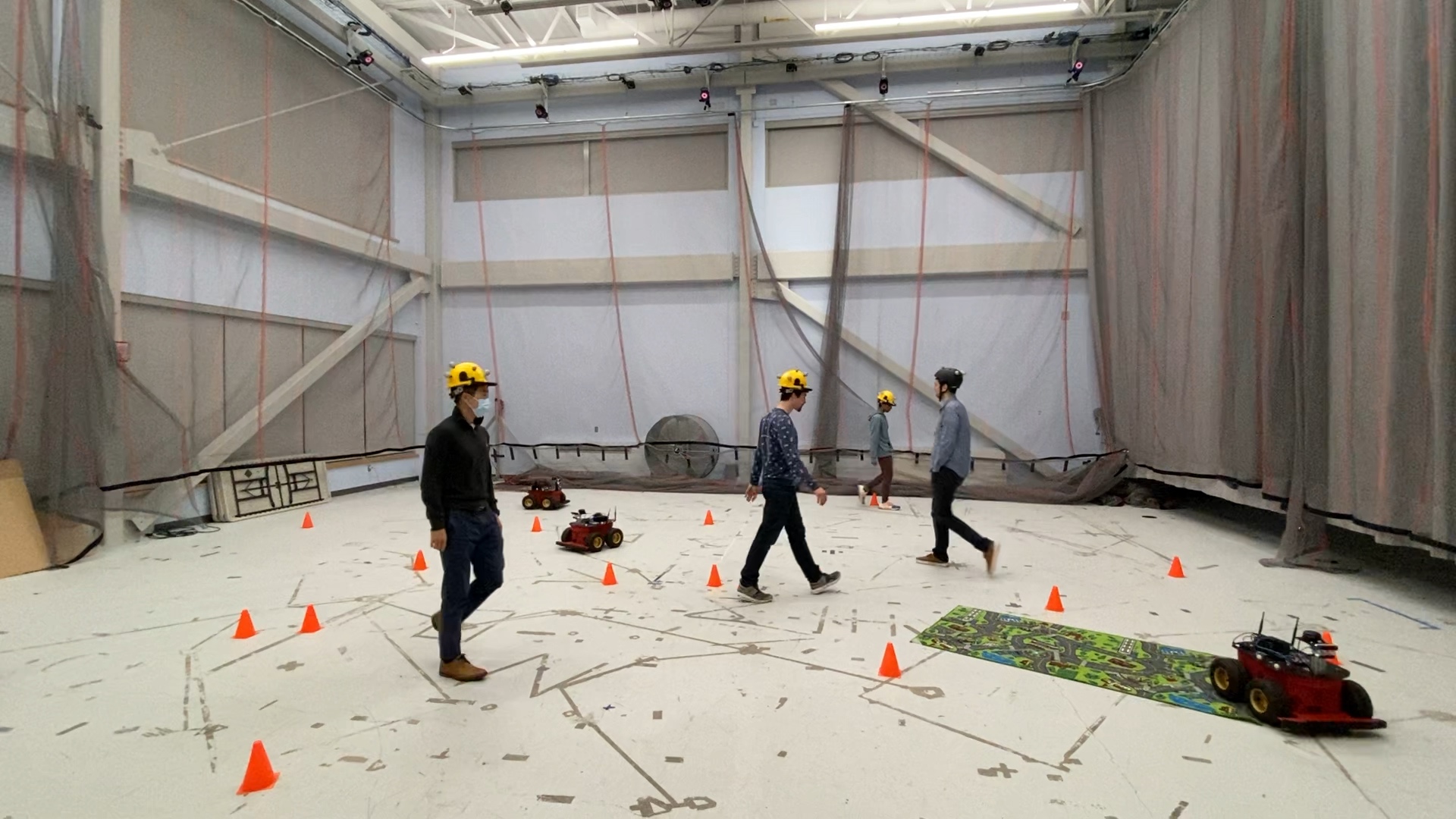}
    \caption{MOT hardware experiment in our motion capture room. We use this setup to demonstrate our MOTLEE algorithm's ability to perform distributed MOT onboard moving robots with localization uncertainty.}
    \label{fig:mot-exp}
\end{figure}

We use MOTA~\cite{bernardin2008clearmot} as the performance metric, given by
\begin{equation}\label{eq:mota}
    \text{MOTA} = 1 - \frac{\Sigma_k(m(k) + fp(k) + mme(k))}{\Sigma_{k}g(k)},
\end{equation}
where $m(k)$ is the number of ground truth objects missed in frame $k$, $fp(k)$ is the number of false positives (i.e., perceived objects that are not found in frame $k$), $mme(k)$ is the number of mismatches, or objects that were reported under a different ID at time $k$ than the ID previously used for that object, and $g(k)$ is the number of ground-truth objects in the scene.
The components of MOTA are highly sensitive to the system's ability to correctly perform data association and report the correct location of objects. 
This makes it a good candidate for measuring the effects of frame misalignment which causes data association errors and corrupts shared object detection measurements. 

\subsection{Effects of Localization Error}

We now demonstrate the performance of MOTLEE's resilience to frame misalignment and localization error.
First, we present our results in the context of a dataset taken with 5 pedestrians walking around our motion capture room with 4 stationary robots located around the perimeter.
We use this dataset to artificially alter the localization error of each robot by initializing robot frame alignments with incorrect $\mathbf{T}_i^j(k_0)$. 
In this way, we isolate the effects of localization error directly, without also introducing other error associated with performing mobile MOT, e.g., poor object detections from rolling-shutter distortion.

To represent the effects of inter-robot frame misalignment, artificial error is introduced into the system by adding a random, constant bias to the robot pose estimates at the start of a run by initializing ${\hat{\mathbf{T}}_i^j(k_0) = \mathbf{T}_\text{error} \mathbf{T}_i^j(k_0) }$,
where $\mathbf{T}_\text{error}$ is composed of random heading error $\theta_\text{error}\sim\mathcal{N}(0,\sigma_\theta)$ and random translation error in the $x$, $y$ plane with magnitude $t_\text{error}\sim\mathcal{N}(0,\sigma_t)$.
We then capture the performance of the system for varying levels of $\sigma_t$ and $\sigma_\theta$ representing varying levels of localization uncertainty.
We correlate the standard deviations $\sigma_\theta$ and $\sigma_t$ at a ratio of \SI{8.12}{\deg} heading error per \SI{1}{\meter} translation error. 
This ratio is the amount of heading error that would produce \SI{1}{\meter} of error in the estimated position of a tracked object that is a distance of \SI{7}{\meter} away from the robot, the approximate average distance between tracked objects and each robot in this dataset. To produce the following results, we run our full dataset through our MOT framework over 5 different samples at each uncertainty level to represent the average performance of the system as error is introduced.

\begin{figure}[t]
    \centering
    \includegraphics[width=\columnwidth]{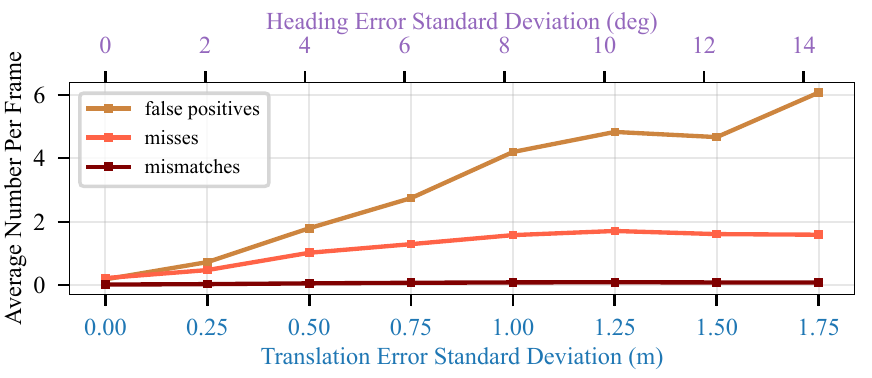}
    \caption{
    Performance of distributed MOT from Casao et al.~\cite{casao2021distributed} under varying levels of artificially introduced localization uncertainty.
    Each of the plotted values directly affects tracking accuracy as defined by MOTA (see~\eqref{eq:mota}).
    Without accounting for localization uncertainty, tracking accuracy degrades as error increases.
    Misses and particularly false positives are the dominating part of the degradation in performance.
    False positive tracks occur because each robot within the multi-view system shares incorrect information about tracked objects, causing the network to report many different tracks of objects that do not exist.
    }
    \label{fig:fp_fn_mme}
\end{figure}

\begin{figure}[t]
    \centering
    \includegraphics[width=\columnwidth]{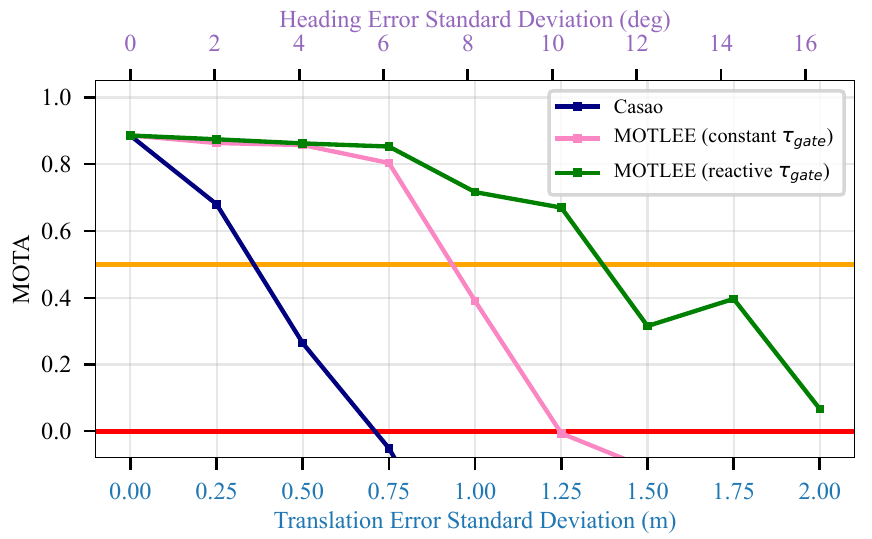}
    \caption{MOT performance results as localization error is introduced. Two lines  at MOTA = 0.5 and MOTA = 0.0 are given as references for understanding the performance degradation. 
    An example scenario that would earn a MOTA of 0.5 is tracking only half of the visible objects and missing the other half. 
    For a MOTA of 0.0, a system could similarly miss half of all visible objects but additionally predict that each of those objects are in some other location.
    MOT performance rapidly degrades with the introduction of localization error (navy line). Performing realignment with dynamic object tracks (pink line) realigns frames successfully in low-uncertainty regimes,
    and performing realignment with a reactive data association tolerance $\tau_\text{gate}$ (green line) makes the system robust to even greater amounts of error. }
    \label{fig:mota-static-results}
\end{figure}

Fig.~\ref{fig:fp_fn_mme} and Fig.~\ref{fig:mota-static-results} show studies on the isolated effects of localization error.
Fig.~\ref{fig:fp_fn_mme} decomposes the different pieces of MOTA and shows how misses and particularly false positives break a system's MOT performance as localization error is introduced.
Fig.~\ref{fig:mota-static-results} demonstrates that while Casao~et~al.'s~\cite{casao2021distributed} MOT algorithm rapidly breaks down with frame alignment error in the system, our MOTLEE algorithm is robust to localization error. 
Frame realignment is performed here by only aligning detections of dynamic objects.

\subsection{Mobile Experiment}

In this section, we evaluate MOTLEE using a team of three mobile ground robots moving along pre-determined trajectories while four pedestrians walk among the robots.
By nature of the small room that the experiment is run in, each robot follows a circular path in its assigned region of the room.
Because of this, cameras have limited co-visibility of objects for approximately half of the time while they face the walls.
In this experiment, there are too few shared dynamic detections to be used for frame realignment (${\eta < \tau_\eta}$), so static landmarks are used to realign frames.
These landmarks are detected and mapped by each robot using the L515 camera.
Detections of cones are made using YOLOv7~\cite{wang2022yolov7} with weights from~\cite{coneslayer}.
The corresponding 3D points within the 2D bounding box are then identified by color thresholding and a 3D cone position is estimated using the median of these points.
Cone detections are accumulated into local maps using the robot's odometry combined with 3D cone detections to chain pairwise associations.
Local maps are shared with neighbors at a frequency of \SI{1}{\hertz}, and frame alignment is performed using these local maps as they arrive.
We represent the frame alignment uncertainty, $\mathbf{\Sigma}^j_i$, as a diagonal covariance matrix with elements determined by a linear scale of the difference between the current and most recent frame alignments.
For this experiment we chose to set $\tau_\eta = $ 100 recent dynamic object detections and $\tau_\text{gate} = 2.0$.

\begin{figure}[t]
    \centering
    \includegraphics[width=\columnwidth]{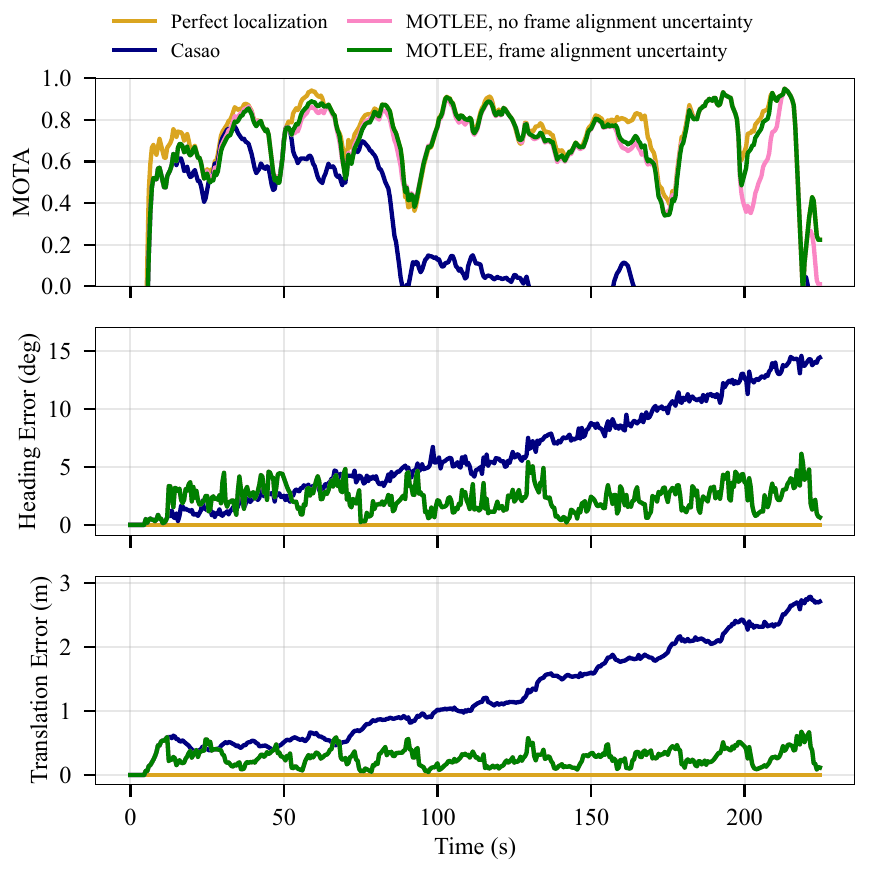}
    \caption{Four-minute long mobile MOT test results.
    MOTA is usually normalized by the sum of the number of objects in each frame over a whole run. 
    However, to show the evolution of tracking over time, the MOTA score is shown over a sliding window of 10 seconds. Casao~et~al.'s algorithm~\cite{casao2021distributed}, which does nothing to correct for localization error, quickly degrades as drift accumulates in the robots. 
    We show that MOTLEE demonstrates improved robustness to frame misalignment and is able to achieve results similar to those of a system with ground truth localization. We also show that by incorporating frame alignment uncertainty into exchanged information in the KCF (green line), MOTLEE achieves a higher MOTA than when this uncertainty is ignored (pink line) even though frame alignment results do not change (the pink and green lines are on top of each other in the heading and translation error plots). }
    \label{fig:mobile-results}
\end{figure}

Fig.~\ref{fig:mobile-results} shows that using only noisy odometry readings without realignment only works well while localization error in the experiment is small, which only occurs at the beginning of the run.
However, as more error is accumulated in each of the robot's frame alignments, $\hat{\mathbf{T}}_i^j(k)$, the system fails to perform collaborative object tracking accurately. 
In contrast, when using our MOTLEE framework for realignment, we are able to perform near the level of the system using ground truth localization.
We achieve an average MOTA of $0.724$, similar to the ground truth performance of $0.756$, while Casao~et~al.~\cite{casao2021distributed} breaks down due to the static camera assumption and scores a MOTA of $0.141$.

\begin{figure}[t]
    \centering
    \includegraphics[width=\columnwidth]{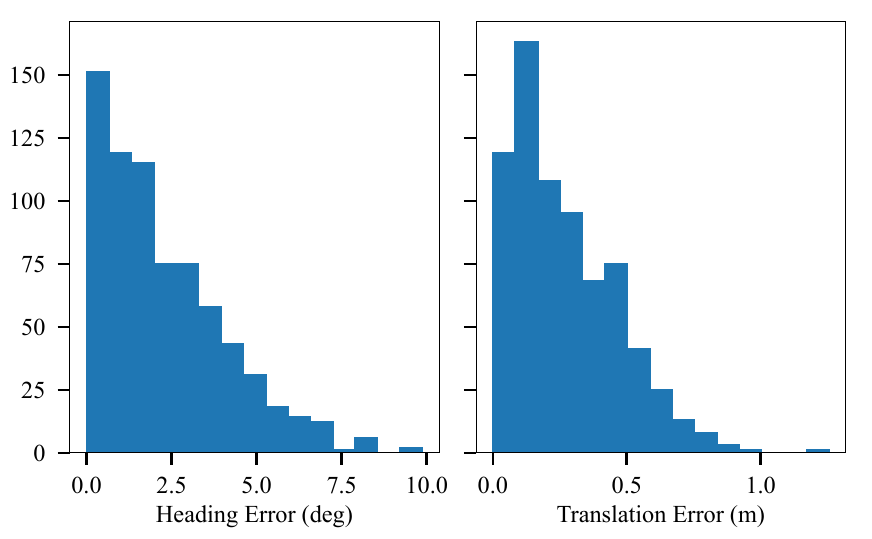}
    \caption{Histogram of error in frame realignment estimates during mobile MOT experiment.}
    \label{fig:error-hist}
\end{figure}

We also show the error between ${\mathbf{T}}_i^j(k)$ and $\hat{\mathbf{T}}_i^j(k)$ in Fig.~\ref{fig:error-hist}.
With our MOTLEE frame realignment, we get a median error of \SI{1.83}{\deg} heading error and \SI{0.22}{\meter} translation error.
Although we use the recorded data to run the system offline so that we can compare different localization scenarios, our algorithm can easily be run in real-time with each tracking and mapping cycle taking an average of \SI{7.1}{\ms} with a standard deviation of \SI{2.9}{\ms} and each frame alignment cycle taking an average of \SI{76.7}{\ms} with a standard deviation of \SI{23.9}{\ms}.

%% file: paper/06_conclusion.tex
\section{Conclusion}\label{sec:conclusion}

This work presents MOTLEE, a distributed MOT method that eliminates the effects of localization error on frame alignment between robots. 
We give methods for realigning frames between pairs of robots using dynamic object detections and static landmarks.
Additionally, we show how to incorporate the uncertainty of localization and of frame alignment into the KCF.
Using these methods, we show through hardware experiments that MOTLEE is robust to localization errors and is able to perform MOT at levels similar to a system with perfect localization.
Future work includes removing the neccessity of an initial alignment (e.g., using \cite{lusk2021clipper}), incorporating frame alignment updates with a filter, integrating collaborative SLAM into MOTLEE, and running MOTLEE online and in real time.